\documentclass[conference]{IEEEtran}
\IEEEoverridecommandlockouts
\usepackage{cite}
\usepackage{amsmath,amssymb,amsfonts}
\usepackage{algorithmic}
\usepackage{graphicx}
\usepackage{textcomp}
\usepackage{xcolor}
\usepackage{float} 
\usepackage{url}
\usepackage{fancyhdr}

\newcommand{\linebreakand}{%
    \end{@IEEEauthorhalign}
    \hfill\mbox{}\par
    \mbox{}\hfill\begin{@IEEEauthorhalign}
}

\makeatother

\def\BibTeX{{\rm B\kern-.05em{\sc i\kern-.025em b}\kern-.08em
    T\kern-.1667em\lower.7ex\hbox{E}\kern-.125emX}}

\fancypagestyle{firstpage}{
    \fancyhf{}
    
    \fancyhead[L]{\small \textcolor{red} {This paper has been accepted in 4th OPJU International Technology Conference (IEEE OTCON 4.0) on Smart Computing for Innovation and Advancement in Industry 5.0. Published version of the paper will be available in IEEE Xplore very soon.}}
    \fancyhead[C]{}
    \fancyhead[R]{}

}

\begin{document}

\title{Find Matching Faces Based On Face Parameters\\


}

\author{\IEEEauthorblockN{Setu A. Bhatt}
\IEEEauthorblockA{\textit{Department of Information Technology } \\
\textit{Dharmsinh Desai University }\\
Nadiad, India\\
bansetu1@gmail.com}
\and
\IEEEauthorblockN{Prof. (Dr.) Harshadkumar B. Prajapati}
\IEEEauthorblockA{\textit{Department of Information Technology} \\
\textit{Dharmsinh Desai University }\\
Nadiad, India \\
prajapatihb.it@ddu.ac.in}
\and
\IEEEauthorblockN{Prof. (Dr.) Vipul K. Dabhi}
\IEEEauthorblockA{\textit{Department of Information Technology } \\
\textit{Dharmsinh Desai University }\\
Nadiad, India\\
vipuldabhi.it@ddu.ac.in}
\and
\IEEEauthorblockN{Ankush Tyagi}
\IEEEauthorblockA{\textit{Software Development Manager} \\
\textit{Ericsson}\\
Austin, Texas, USA\\
ankush.tyagi@ericsson.com}
}

\maketitle

\thispagestyle{firstpage}

\begin{abstract}
This paper presents an innovative approach that enables the user to find matching faces based on the user-selected face parameters. Through gradio-based user interface, the users can interactively select the face parameters they want in their desired partner. These user-selected face parameters are transformed into a text prompt which is used by the Text-To-Image generation model to generate a realistic face image. Further, the generated image along with the images downloaded from the Jeevansathi.com are processed through face detection and feature extraction model, which results in high dimensional vector embedding of 512 dimensions. The vector embeddings generated from the downloaded images are stored into vector database. Now, the similarity search is carried out between the vector embedding of generated image and the stored vector embeddings. As a result, it displays the top five similar faces based on the user-selected face parameters. This contribution holds a significant potential to turn into a high-quality personalized face matching tool.
\end{abstract}


\section{Introduction}
Finding a partner that aligns with one’s individual preferences is one of the greatest ambition in one’s life. To achieve this, there are various matrimonial sites currently present in the market, be it a traditional or AI driven matrimonial sites. In traditional matrimonial sites, such as Jeevansathi.com and Shaadi.com, it enables the individuals to connect through variations, such as, personality, taste, and background. These platforms enables the users to enter their partner’s preference details like age, religion, education, profession, marital status and interests. Later on, they utilise these criteria to match the profiles and find ideal partners for the individual. After the traditional matrimonial sites, AI driven matchmaking platforms like betterhalf.ai is introduced which focuses on six dimensions of personality that includes relationships, social, intellectual, emotional, values and physical levels. Profiles are matched on the basis of individual compatibility score as well as overall compatibility score \cite{1}.

Currently, the algorithms used in such sites are only limited to the textual data processing, which means they cannot include visual opinions about their partner. Although the facial recognition systems are currently evolving at huge pace, many sites still do not allow the users to choose a partner by adjusting face parameters. Researchers in \cite{2} conducted a study on matrimonial sites where they concluded that users are dissatisfied with the services provided by the matrimonial sites. Therefore, it is obvious that there is a need for an interactive and customized face-matching system where the users can specify the face parameters, they consider more important for the matching.

The proposed system leverages the combination of deep learning models along with the capabilities of vector database to retrieve the similar faces based on the face parameters selected by the user. It is comprised of various components starting from the image generation, followed by face detection, feature extraction and similarity search. Since the similarity search will be performed between the generated image and the images present in our database, so the image generation plays a crucial role in this proposed system. Therefore, it is necessary to provide clear, precise and accurate prompt to the Text-To-Image generation model to generate the image accordingly. Further, the user can only specify the face parameters, and not create the prompt in order to prevent any misuse of the system. Face detection model will detect the facial region from the image, and the feature extraction model will extract the facial features, including eyes, nose, lips and others from the detected facial region and transform it into vector embedding of 512 dimensions. The use of vector database helps to carry out faster similarity search between the embeddings.

The system is completely customizable and can be tailored according to any use case. The only change that one need to do is to store images related to your use case into the database. This system has immense potential in the domain like (1) Film Industry and (2) Police Department. In the film industry, suppose the screenwriter has written the script for a film. According to the script, the actor needs to have certain face parameters in accordance to match with the role offered in the film, e.g. the actor needs to have almond shaped eyes, pointed nose, and others. So, the casting director will select the face parameters according to the role, and based on it, our proposed system returns which actors will be most suitable for this role, assuming that the database contains the images of the actors. 
If a crime happens in the resident areas with no security surveillance systems then it is difficulty for the police department to findout the criminal. Generally, the victim will describe the person who has done the crime. In earlier days, the police used to call the sketch artist to prepare the sketch of the face based on the description provided by the victim. Now, if the database contains the Aadhaar card images of all the citizens in a country, or the residents of particular area, then we can retrieve the faces that matches with the image generated based on the description provided. This may help the police to find the criminal. 

The remainder of this paper is structured as follows: Section II reviews the related work and existing matchmaking platforms. Section III discusses the proposed methodology. Section IV demonstrates the experimental results obtained from the important components. Section V concludes the paper and suggests directions for the future work.

\section{Related Work}


There are many matrimonial sites, such as, Jeevansathi.com, Shaadi.com, Bharat Matrimony, Betterhalf.ai, and many more present in the market. In traditional matrimonial sites, the main focus is to find potential matches on the basis of criteria specified in the user’s partner preference details. At the time of registration, these sites collect various details about the user as well as the potential partner preference of the user. Several details are collected from the user and are fed into predefined programs that evaluate the ideal matches for the individuals. These ideal matches closely align with the textual preference provided by the user. A user can search for another user by providing 'some criteria' or providing the 'exact profile ID' of the user. But, ever wondered what happens when the user enters incorrect details or creates a fake profile?

Moving further, AI-powered matrimonial sites like betterhalf.ai use the power of AI along with machine learning algorithms to find the ideal partners for the individuals. The user is smoothly onboarded by entering necessary details, such as name, age, and gender. Based on these details, they determine the religion, mother tongue, and some other relevant details. Using the IP address, the current location of the user is identified. These collected data along with the predicted data are provided to the AI and machine learning driven match compatibility algorithms along with the behavioral data to predict the ideal matches for the user. Basically, it analyzes and matches the user preference and behavior. It adopts the concept of a feedback loop to refine the matches provided by the recommendation algorithm. For every predicted match, the user is presented with two options i.e. Pass or Interested to determine whether the displayed match is relevant to them or not. Suppose that the user selects ‘pass’, then the recommendation algorithm reduces the possibility of predicting similar matches in the future. And if the user selects ‘interested’, then the recommendation algorithm increases the possibility of predicting similar matches with specific attributes. Researchers \cite{3} introduced an AI-driven, NLP-based matchmaking system which aims to create a personalized matchmaking system, still it has the limitation of use of textual data only. The work presented in this system closely aligns with the approach used in \cite{4} and future work suggested in \cite{3}. 

Researchers in \cite{2} have also concluded that users are more dissatisfied than satisfied by the services provided by these matrimonial sites. Therefore, there is a need for the personalised matchmaking system which enhances more user engagement and satisfaction. The highest priority should be given to the face of potential partner perceived by the user. In almost every matrimonial site, there is no option to specify how the face of their desired partner should look like. This proposed system helps to address the limitations of current matrimonial sites. This system lets the user specify the face parameters they desire in their potential partner. Based on these selected face parameters, a face incorporating all the user-selected face parameters is generated, which is further used for the similarity search with the faces present in our database.

\section{Proposed Work}

\begin{figure}[hptb]
    \centerline{\includegraphics[scale=0.35]{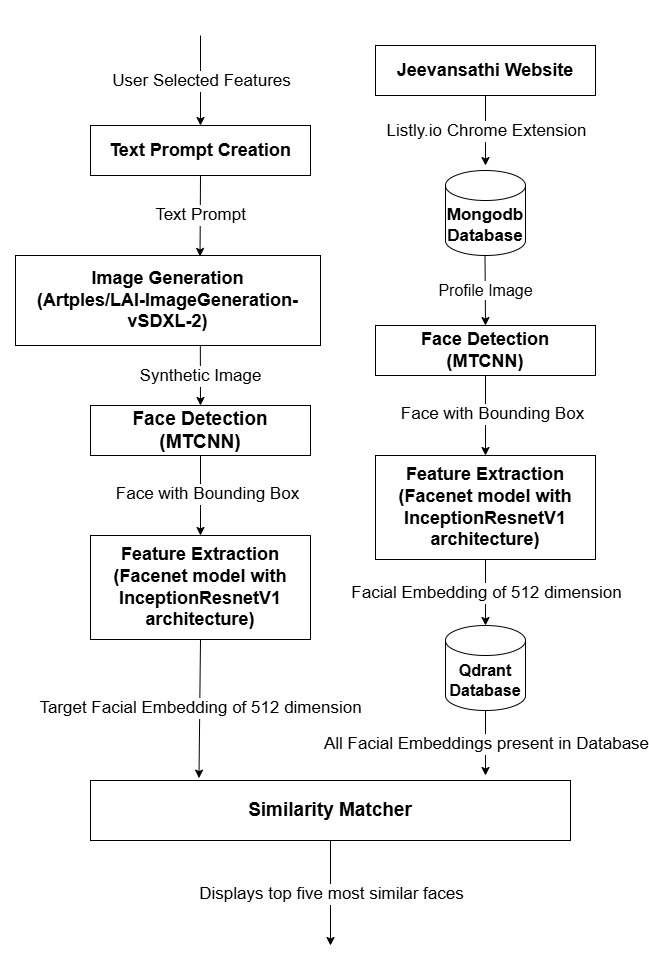}}
    \caption{Architecture of the proposed system}
    \label{fig:architecture}
\end{figure}

This proposed system is user preference driven which facilitates the user to find matching faces based on the user-selected face parameters. Figure \ref{fig:architecture} illustrates the architecture of proposed system which consist of various pretrained deep learning models, external API and vector database to enhance the overall performance of the system.

\subsection{Overview of proposed system architecture}

The proposed system can be divided into a three-tier architecture which includes the presentation layer, application logic and data layer. The presentation layer, a user-facing component, is built using Gradio, a python library, that interacts with the user allowing it to test the working of model effectively through an user friendly interface. Firstly, it allows the user to select the face parameters, including eyes shape, nose shape, eyebrow shape, and others. Further based on these selected face parameters, it displays the generated text prompt, image and similar images. The application logic, the core processing part of entire system, is built using the python environment in the Google Colab that executes the core logic of this proposed system, starting from text prompt creation to similarity search. Based on the user input, subsequent actions are fired that handles various tasks like image generation, face detection, feature extraction and similarity search. For the data storage and retrieval, we have used combination of non-relational database and vector database. The non-relational database, MongoDB, was used to store the images downloaded from Jeevansathi.com \cite{10}. While, the vector database, Qdrant, was used to store the feature vector that are efficient for similarity searches.

\subsection{Data Collection and Storage}
The Listly.io, was used for downloading the data of Jeevansathi.com \cite{10}. It displays 91 ‘my matches’ everyday along with ‘top-picked matches’ for the premium user. So, the gathered data is comprised particularly of image url and name of the person. Now, some preprocessing is carried out on the scraped data, such as removing null image urls.

\subsection{Defining Face Parameters and Text Prompt Creation}
Defining the face parameters is the most crucial step, as it is the foundational block of our entire system.
\begin{enumerate}
    \item The users can specify the shape of face parameters, including eyes, eyebrows, nose, lips, jawline, chin, face, beard and moustache. The users can also specify additional information like the gender, age group and skin tone.
    \item These parameters are used for the generation of text prompt. The generated text prompt is given as an input to the Text-to-Image generation model. The final image generated by the model will be as per the user-specified face parameters.    
\end{enumerate}

\subsection{Image Generation Model}
Text-To-Image generation model is one of the core component of our proposed system. The entire system depends on the quality of image generated. After trying out multiple models from the Hugging Face, the output produced by model “Artples/LAI-ImageGeneration-vSDXL-2” outperforms all the other models in generating realistic images. This model uses “stable-diffusion-xl-base-1.0” from the “stabilityai” organization as the base model, where it generates noisy latent using base model and further refines them using a refinement model. This model works as follow:
\begin{enumerate}
    \item The generated text prompt is passed as input to the model. \item The model is inferenced using the Hugging Face inference api, containing the headers along with the generated text prompt. 
    \item The realistic image is generated which incorporates all the user-specified face parameters.
\end{enumerate}

\subsection{Face Detection}
Detecting the facial area from an image itself comes up with various challenges in our proposed system, such as, presence of multiple faces in an image, different angles of the face, presence of face accessories like sunglasses and others \cite{5}. The data downloaded from Jeevansathi.com exhibits all the above mentioned constraints. To accomplish this task of face detection and facial landmark detection, various models were used. The models indulged in order to accomplish it were Multi-task cascaded neural network and Dlib’s 68 facial landmark detection. The Multi-task Cascaded Convolutional Neural Network works as follow:
\begin{enumerate}
    \item Initialise MTCNN model with the parameter keep-all set to false meaning that it detects the single largest face from the image, else it will detect all the faces present in the image. So, with the help of specified parameter, the problem of presence of multiple faces in an image gets solved.
    \item It finds the face in the provided input image, and returns the bounding box coordinates for the detected face. The OpenCV library takes the bounding box coordinates and draws rectangle with red colour of thickness 2 around the detected face region.
\end{enumerate}

\subsection{Feature Extraction and Vector Representation}
The most important step of our proposed system is to extract the features from the face and map the extracted features from the detected face into a high dimensional vector embedding of 512 dimension. To accomplish this task, the power of FaceNet model with InceptionResnetV1 architecture is leveraged. InceptionResnetV1 is used as Feature Extractor, while FaceNet is used to map the extracted features into high dimensional vector embedding. The InceptionResnetV1 leverages the power of Inception Module to extract features at various scale, while residual connections are utilised to prevent the effect of vanishing gradient. The FaceNet employs triplet loss function, which consists of three images: anchor image (original image), positive image (another image of same person) and negative image (image of different person)\cite{6}. 

\subsection{Similarity Matcher}
The vector embedding of 512 dimensions are stored into vector database. The similarity search is performed with the capabilities of vector database. This works as follow:
\begin{enumerate}
    \item We need to specify the target vector embedding generated from the generated image, and the number of similar images you want to retrieve.
    \item Initialize a Qdrant client, provide the collection name along with the above mentioned details.
\end{enumerate} 

\section{Experiment Results}

\subsection{User Feature Selection and Text Prompt Creation}

\begin{figure}[hptb]
    \centerline{\includegraphics[scale=0.45]{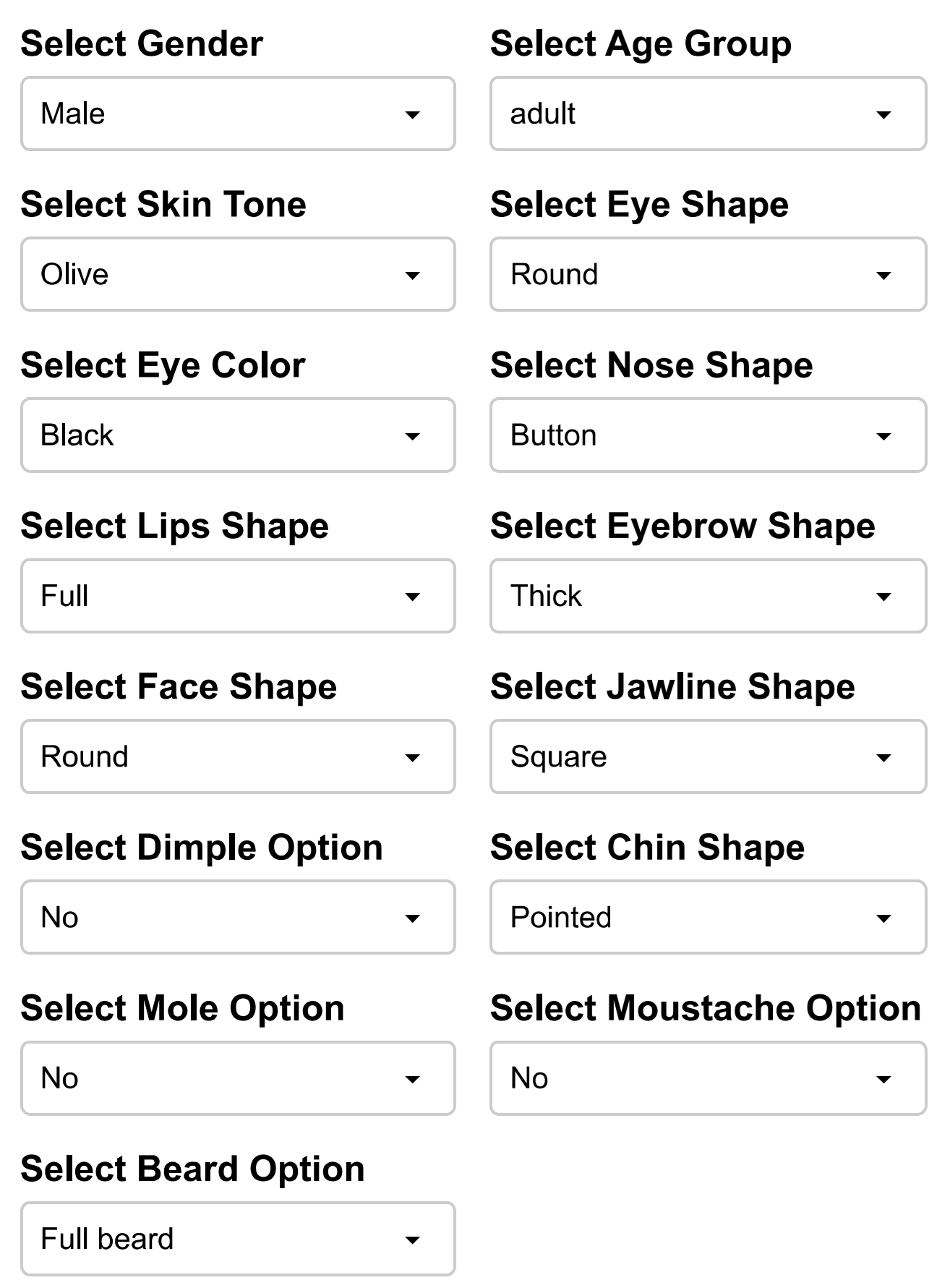}}
    \caption{Gradio user interface for selection of face parameters by the user}
    \label{fig:GradioUI}
\end{figure}

Figure \ref{fig:GradioUI} displays the face parameters selected by the user. The user is looking for an adult male with olive skin tone, round eye shape with black eyes, button nose, and full lips, thick eyebrows, round face shape, square jawline, pointed chin, and a full beard.

\begin{figure}[hptb]
    \centerline{\includegraphics[width=0.8\linewidth]{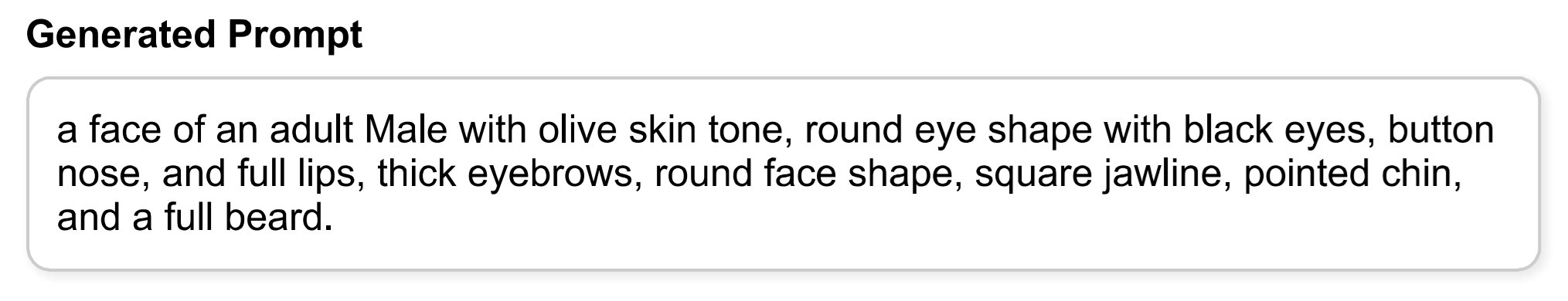}}
    \caption{Text prompt generated for the user-selected face parameters in Fig. \ref{fig:GradioUI}}
    \label{fig:Textprompt}
\end{figure}

Figure \ref{fig:Textprompt} presents the text prompt generated based on the face parameters selected by the user in Figure \ref{fig:GradioUI}\\

\subsection{Image Generation}
Fig. 4 presents the result of Text-To-Image generation model used in our proposed system. This image is generated based on the text prompt depicted in Fig. 3. The image looks very realistic.

\begin{figure}[hptb]
    \centerline{\includegraphics[scale=0.6]{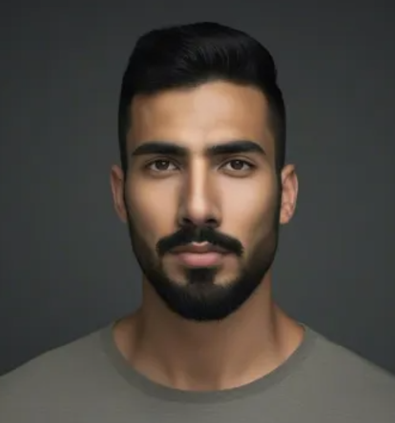}}
    \caption{Image generated based on the text prompt generated in Fig. 3}
    \label{figure 3}
\end{figure}

As discussed earlier that several Text-To-Image generation model were used to generate the face images. Here are some of the results of images produced by the models. The prompt given to these models was “a face of an adult Female with fair skin tone, almond-shaped eye shape with black eyes, button nose, and full lips, straight eyebrows, oval face shape, square jawline, pointed chin”.
\begin{figure}[hptb]
    \centerline{\includegraphics[scale=0.5]{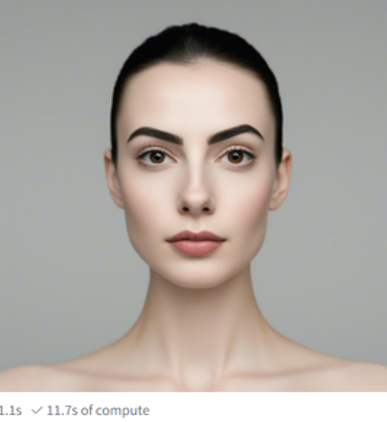}}
    \caption{The result generated by Artples/LAI-ImageGeneration-vSDXL-2 from Hugging Face with compute time of 11.7 seconds.}
    \label{figure 5}
\end{figure}
\begin{figure}[hptb]
    \centerline{\includegraphics[scale=0.5]{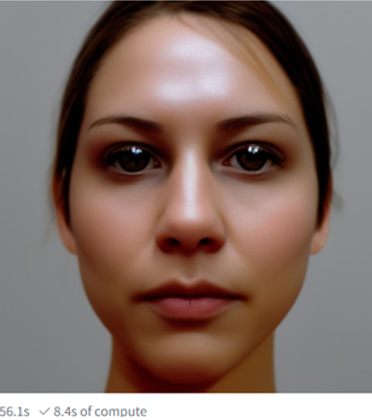}}
    \caption{The result generated by stabilityai/stable-diffusion-2 from the Hugging Face with compute time of 8.4 seconds.}
    \label{figure 6}
\end{figure}
\begin{figure}[hptb]
    \centerline{\includegraphics[scale=0.5]{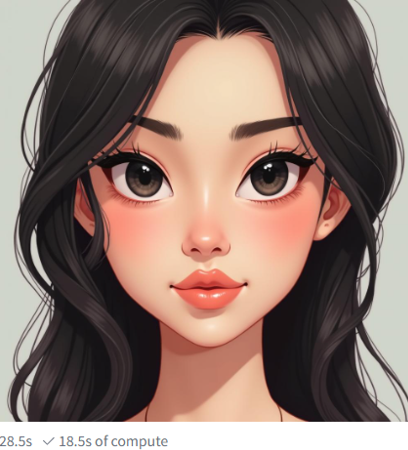}}
    \caption{The result generated by black-forest-labs/FLUX.1-dev from Hugging Face with compute time of 18.5 seconds}
    \label{figure 7}
\end{figure}

Fig. 5, 6, 7 displays the result generated by the Artples/LAI-ImageGeneration-vSDXL-2, stabilityai/stable-diffusion-2 and black-forest-labs/FLUX.1-dev models respectively. From the results, we can clearly see that the results produced by the Artples/LAI-ImageGeneration-vSDXL-2s model has outperformed the results produced by the others. The model produced modern as well as realistic image.

\begin{figure}[hptb]
    \centerline{\includegraphics[scale=0.5]{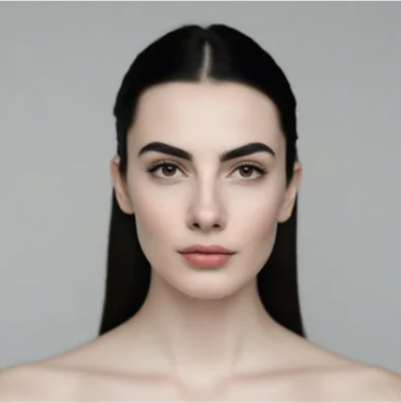}}
    \caption{The image generated for the female}
    \label{figure 8}
\end{figure}

Fig. 8 presents the image generated for the female when the text prompt provided was “a face of adult Female with fair skin tone, almond-shaped eye shape with black eyes, straight nose, and full lips, thick eyebrows, oval face shape, square jawline, pointed chin”.
\subsection{Face Detection and Feature Extraction}
Fig. 9 presents the face localization when multiple faces are present in the image using multi-task cascaded neural network. It selects the single largest face as output.
\begin{figure}[hptb]
    \centerline{\includegraphics[scale=0.5]{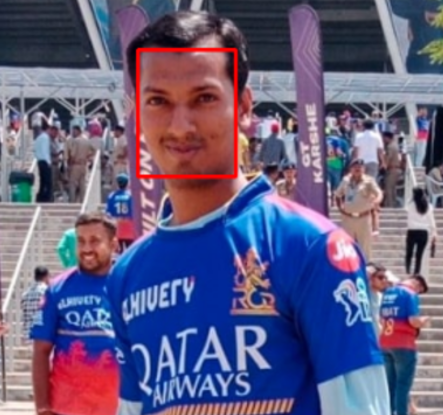}}
    \caption{Face detection from image with multiple faces }
    \label{figure 9}
\end{figure}

\begin{figure}[hptb]
    \centerline{\includegraphics[scale=0.5]{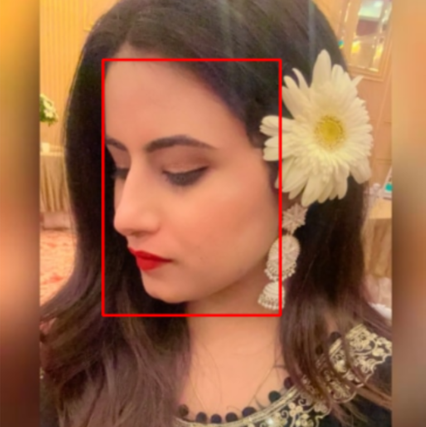}}
    \caption{Face detection in the side profile}
    \label{figure 10}
\end{figure}

Fig. 10 presents the face detected in different angle using multi-task cascaded neural network.

\begin{figure}[hptb]
    \centering
    \begin{tabular}{cc}
        \multicolumn{1}{c}{A} & \multicolumn{1}{c}{B} \\
        \includegraphics[height=0.18\textheight]{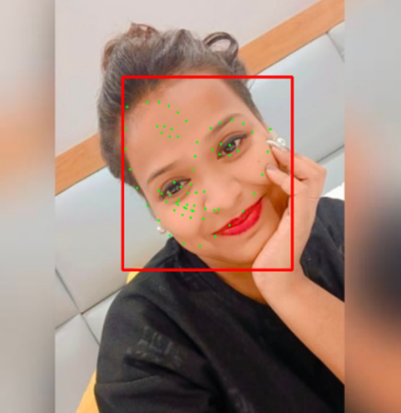} &
        \includegraphics[height=0.18\textheight]{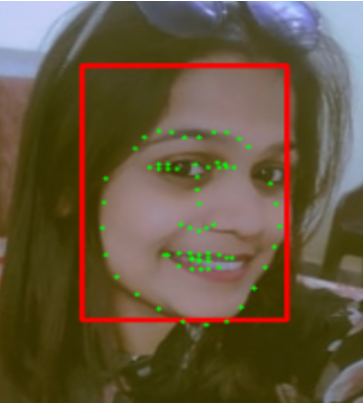}
    \end{tabular}
    \caption{Facial landmark detection using Dlib's Shape Predictor}
    \label{fig:facial_landmark_detection1}
\end{figure}

Fig. 11 (A) and Fig. 11 (B) present the 68 facial landmarks detected using the Dlib’s Shape Predictor. The facial landmarks are not correctly identified. Researchers in \cite{7} and \cite{8} have proven that the Multi-task cascaded neural network produces more accurate results than the Dlib.

\subsection{Similar Images}

\begin{figure}[hptb]
    \centering
    \begin{tabular}{cc}
        \multicolumn{1}{c}{A} & \multicolumn{1}{c}{B} \\
        \includegraphics[height=0.18\textheight]{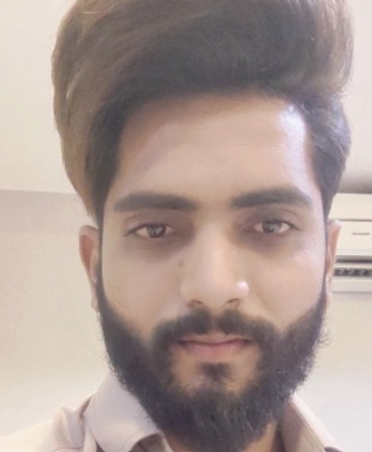} &
        \includegraphics[height=0.18\textheight]{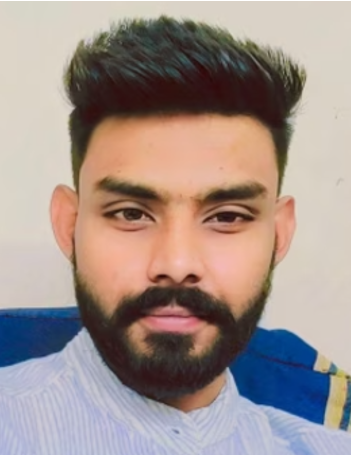}
    \end{tabular}
    \caption{Similar Images}
    \label{fig:facial_landmark_detection2}
\end{figure}

Fig. 12 (A) presents the most similar image and Fig. 12 (B) presents the other similar images that closely aligns with the face presented in the Fig. 4.
\begin{figure}[hptb]
    \centerline{\includegraphics[scale=0.55]{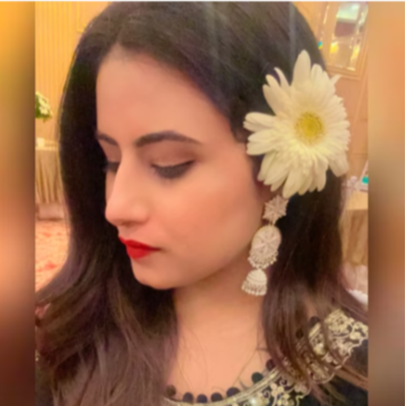}}
    \caption{Most similar face that closely aligns with face presented in Fig. 8}
    \label{figure 12}
\end{figure}

Fig. 13 presents the most similar image that closely aligns with the face presented in Fig. 8.
\section{Conclusion and Future Work}
The main goal of the proposed system was to find matching faces based on the user-selected face parameters. This system acts as a bridge between the advanced face recognition technologies and the current matchmaking platforms. Current matrimonial platforms are limited to textual data processing. These platforms utilize the user’s partner preference details to find the ideal match for them. The proposed system leverages the cutting-edge power of deep learning models, capabilities of Text-To-Image generation model, and user-preference to find the faces that closely aligns with the face of their desired partner. The system is flexible to be used across various domains, like film industry or police department. The only requirement is to change the dataset as per the use case. The system holds the significant potential to turn into high quality personalized face matching tool in the future.

The images generated via Text-To-Image generation model were realistic in nature. The model can generate different faces for different shapes of eyes, eyebrows, nose, lip, beard and moustache. Although the proposed system demonstrates the remarkable potential in finding matching faces based on user-selected face parameters, yet there are certain limitations that need to be addressed for the broader applicability. One of the limitation lies in generating the image with special face parameters like mole or dimple. The proposed approach of finding matching faces based on user-driven face parameter selection has introduced the possibility for future advancements in the fields such as matchmaking and security, where personalized face matching is crucial.

\bibliographystyle{ieeetr}
\bibliography{refs}

\end{document}